\documentclass[letterpaper, 10 pt, journal, twoside]{IEEEtran} 

\usepackage{graphicx} %
\usepackage{epsfig} %
\usepackage{mathptmx} %
\usepackage{times} %
\usepackage{amsmath} %
\usepackage[dvipsnames]{xcolor}
\usepackage{duckuments}
\usepackage{amssymb}  %
\makeatletter
    \let\NAT@parse\undefined
\makeatother
\usepackage[square,numbers,sort&compress]{natbib}
\usepackage{cuted}
\usepackage{caption}
\usepackage{booktabs}
\usepackage{multicol}
\usepackage{multirow}
\usepackage{wrapfig}

\usepackage{enumitem}
\usepackage{algorithm}
\usepackage{spverbatim}
\usepackage[noend]{algpseudocode}

\newcommand{\rebuttal}[1]{{{#1}}}
\newcommand{\camready}[1]{{{#1}}}
\newcommand{\ralparagraph}[1]{{\bf #1.}}

\usepackage{hyperref}
\usepackage{hypcap} %
\usepackage{subcaption}

\hypersetup{
    colorlinks=true,
    linkcolor=MidnightBlue,
    filecolor=magenta,      
    urlcolor=MidnightBlue,
    citecolor=MidnightBlue,
}
\usepackage[capitalise, nameinlink]{cleveref}
\title{
CuriousBot: Interactive Mobile Exploration\\
via Actionable 3D Relational Object Graph
}

\author{Yixuan Wang$^{1,2}$, Leonor Fermoselle$^{2}$, Tarik Kelestemur$^{2}$, Jiuguang Wang$^{2}$, and Yunzhu Li$^{1}$
\thanks{Manuscript received: October 1, 2025; Revised: January
3, 2026; Accepted: February 2, 2026.}
\thanks{This paper was recommended for publication by Editor Markus Vincze upon evaluation of the Associate Editor and Reviewers' comments.
This work was supported in part by the DARPA TIAMAT program (HR0011-24-9-0430) and the Toyota Research Institute (TRI).
}
\thanks{$^{1}$Yixuan Wang and Yunzhu Li are with School of Engineering and Applied Science,
 Columbia University, USA (email: yixuan@cs.columbia.edu; yunzhu.li@columbia.edu).}
\thanks{$^{2}$Yixuan Wang, Leonor Fermoselle, Tarik Kelestemur, and Jiuguang Wang are with Robotics and AI Institute, USA.}
\thanks{Digital Object Identifier (DOI): 10.1109/LRA.2026.3666384}
}

\begin{document}

\maketitle

\vspace{-20pt}
\begin{strip}
    \centering
    \vspace{-105pt}
    \includegraphics[width=\linewidth]{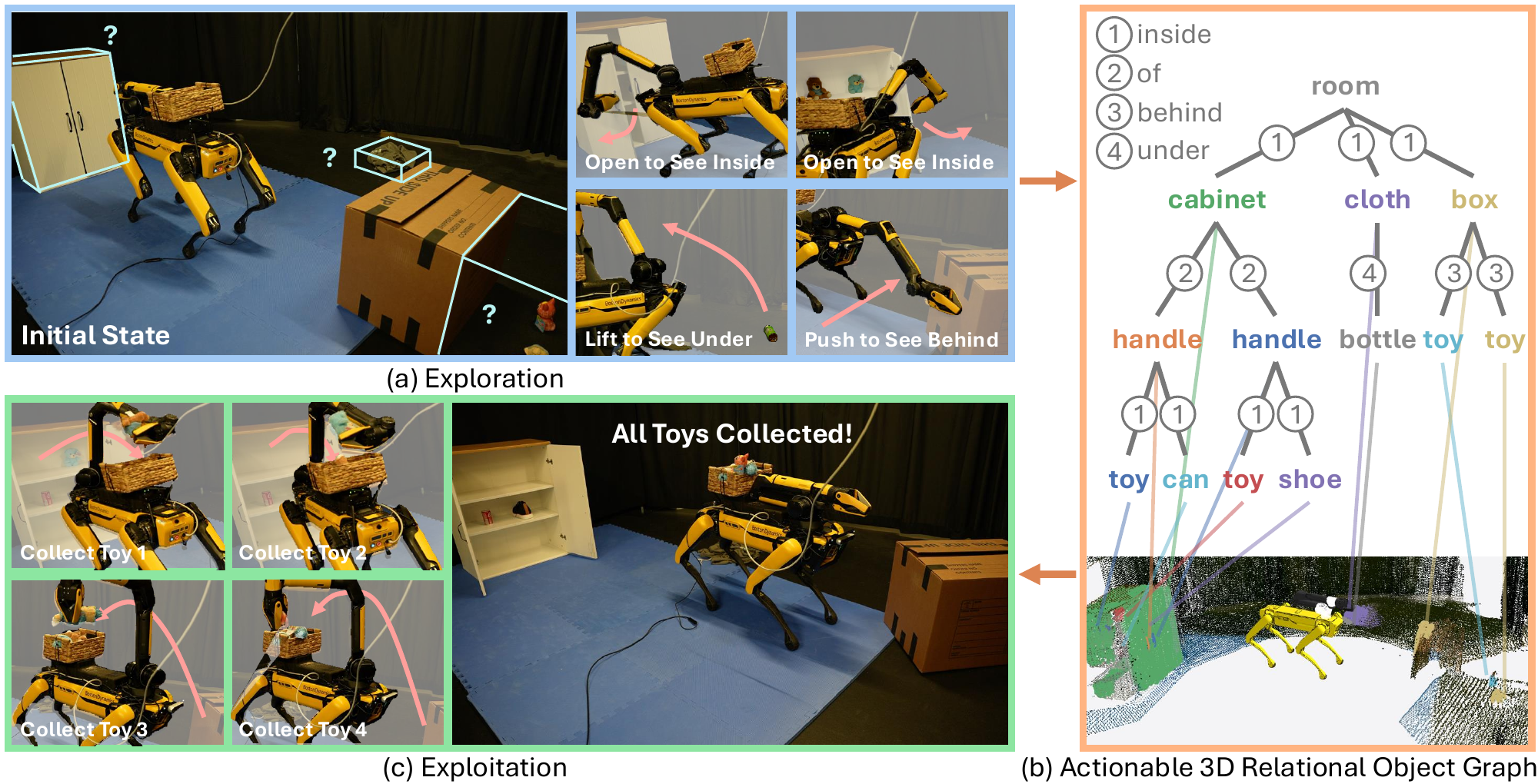}
    \vspace{-18pt}
    \captionof{figure}{\small
    \textbf{CuriousBot.}
    We present a mobile robotic system that can (a) interactively explore the environment, such as inspecting hidden spaces inside a cabinet or behind a box, (b) construct an actionable 3D relational object graph that encodes both the semantic and geometric information of object nodes, along with various object relationships, and (c) perform manipulation tasks by retrieving objects through traversal of the actionable 3D relational object graph.
    }
    \label{fig:teaser}
    \vspace{-15pt}
\end{strip}

\begin{abstract}

Mobile exploration is a longstanding challenge in robotics, yet current methods primarily focus on active perception instead of active interaction, limiting the robot's ability to interact with and fully explore its environment. Existing robotic exploration approaches via active interaction are often restricted to tabletop scenes, neglecting the unique challenges posed by mobile exploration, such as large exploration spaces, complex action spaces, and diverse object relations. In this work, we introduce a 3D relational object graph that encodes diverse object relations and enables exploration through active interaction. We develop a system based on this representation and evaluate it across diverse scenes. Our qualitative and quantitative results demonstrate the system's effectiveness and generalization across object instances, relations, and scenes, outperforming methods solely relying on vision-language models (VLMs). Our project page is available at \href{https://bdaiinstitute.github.io/curiousbot/}{[link]}.

\end{abstract}

\vspace{-13pt}

\section{INTRODUCTION}

\vspace{-5pt}

\IEEEPARstart{E}{xploration} remains a significant challenge for mobile robots, especially in complex household environments filled with occlusions, such as objects concealed within cabinets, hidden under furniture, or obscured behind other obstacles. Traditional exploration methods primarily focus on \emph{active perception}~\cite{cao2021tare, choset2000sensor}, aiming to determine the optimal camera position to minimize unknown spaces, and often neglect the crucial aspect of \emph{active interaction}, which involves deciding where and how to physically interact with the environment to reveal hidden spaces. While recent works such as RoboEXP~\cite{jiang2024roboexp} have considered active interaction, their focus is primarily on tabletop manipulation, limiting their applicability in complex, real-world mobile settings.

In contrast to tabletop scenarios, mobile interactive exploration introduces unique challenges:
\begin{itemize}[leftmargin=10pt]
 \item \textbf{Expanded exploration space}: the exploration area for mobile robots is substantially larger and needs to utilize complex navigation and mapping skills.
 \item \textbf{Complex occlusion relationships}: occlusions in household environments are intricate. While RoboEXP considers basic relationships such as \texttt{on}, \texttt{of}, \texttt{under}, \texttt{behind}, and \texttt{inside}, real-world settings present complex occlusions, such as items hidden beneath furniture or blocked by other objects, requiring more sophisticated reasoning and interaction strategies.
\item \textbf{Larger action space}: mobile exploration involves a broader action space that includes both navigation and manipulation to handle various objects and scenes.
\end{itemize}

In this work, we introduce a system that simultaneously exhibits three unique properties:
1) \textbf{Interactive} – compared to prior mobile object search works~\cite{cao2021tare, choset2000sensor}, which primarily focus on \textit{active perception}, our approach emphasizes \textit{active interaction} with the environment using \textit{diverse} skills.
2) \textbf{Mobile} – compared to RoboEXP~\cite{jiang2024roboexp}, which is limited to tabletop settings and highly constrained mobile scenarios, our system can operate in a much larger space and involve more complex spatial relations, including \texttt{on}, \texttt{of}, \texttt{under}, \texttt{behind}, and \texttt{inside}.
3) \textbf{Exploratory} – compared to prior mobile manipulation works~\cite{jiang2019open, beetz2023cram, garrett2020online}, which cannot reason about occluded space and autonomously discover unknown spaces, our system can actively explore the environment to uncover the unknown spaces.
To our best knowledge, no prior works achieve all three properties simultaneously.

Specifically, we tackle the challenges of active mobile exploration using our 3D relational object graph powered by Visual Foundational Models (VFMs), as summarized in Figure~\ref{fig:teaser}. In contrast, prior object graphs~\cite{honerkamp2024language, Werby-RSS-24} are not action-conditioned and lack occlusion reasoning,
and thus unfit for interactive mobile exploration (e.g., \textit{pushing} a chair to \textit{open} a cabinet). Our system consists of four modules - \textbf{SLAM}, \textbf{Graph Constructor}, \textbf{Task Planner}, and \textbf{Low-Level Skills}, as shown in Figure~\ref{fig:method}.

The SLAM module takes in a sequence of RGBD observations and robot odometry, and outputs the camera pose. Given observations and camera poses, our graph constructor first builds object nodes by detecting and segmenting objects via the open-vocabulary object detector and Segment Anything~\cite{Cheng2024YOLOWorld, kirillov2023segany}.
By leveraging spatial and semantic information, we determine the relationships between nodes, which are then used for downstream task planning. The task planning module takes in the serialized object graph and generates action plans using a Large Language Model (LLM). Finally,  low-level skills, consisting of several action primitives, execute the generated action plan.

We evaluate our system in various scenes requiring exploration and demonstrate its capability to handle a wide range of object categories, including articulated, deformable, and rigid objects. Furthermore, our 3D relational object graph can encode multiple occlusion relationships commonly seen in household environments, such as \texttt{of}, \texttt{on}, \texttt{under}, \texttt{behind}, and \texttt{inside}. The system can adapt to different environment layouts, such as a box-filled room or a living room.
We quantitatively analyze our system by evaluating it across five tasks, each repeated ten times under different initial conditions, and identify common failure patterns.
Additionally, we compare our method with the direct use of GPT-4V to guide robot exploration. Our findings indicate that our 3D relational object graph is more effective for task planning.

In summary, our contributions are threefold: i) We introduce the 3D relational object graph, which can encode a number of common object relations, enabling the mobile robot to explore diverse everyday environments. ii) We develop the CuriousBot system, which can automatically construct the 3D object graph, plan exploration, and interact with the environment to reduce unknown spaces. iii) We conduct comprehensive experiments, demonstrating that our system can fully explore environments and accurately build the object graph. The testing scenes feature diverse object categories, object relations, and scene layouts. Additionally, we provide deeper insights into our system through error breakdown and comparisons with baseline methods.

\vspace{-5pt}
\section{RELATED WORK}
\vspace{-5pt}
\subsection{Robotic Exploration}
\vspace{-3pt}

Robotic exploration is crucial for many applications, including search and rescue~\cite{niroui2019deep, liu2013robotic}, object search
~\cite{zheng2023asystem, yokoyama2024vlfm},
and mobile manipulation~\cite{misra2018mapping, batra2020rearrangement, xia2020relmogen, ehsani2021manipulathor}. The typical objective of exploration is to minimize the unknown areas in the environment
~\cite{cao2021tare, yokoyama2024vlfm, zheng2022towards, zheng2023asystem, chen2019learning}.
However, these methods generally focus on exploration through active perception, neglecting exploration via active interaction, which limits the robot's ability to fully explore environments, such as finding objects inside cabinets.

The work most closely related to ours is RoboEXP~\cite{jiang2024roboexp}, where the robot interacts with the environment to build a complete 3D object graph of the scene. However, their focus is on tabletop scenes, while our mobile setting poses different challenges. In contrast, our approach emphasizes mobile exploration through active interaction, which introduces unique challenges such as larger exploration areas, more complex object relationships, and a broader action space.

Fabian et al. also explore mobile exploration via active interaction~\cite{schmalstieg2023learning}. However, they only consider a subset of the diverse object relationships, which are essential for complex exploration behaviors. On the contrary, our 3D relational object graph can encode five types of object relations. Additionally, they only consider opening skill and rely on AR markers in the real world to guide manipulation. In contrast, we incorporate more skills, including pushing, opening, lifting, flipping, and more, without requiring additional markers.

\begin{figure*}[t]
    \centering
    \includegraphics[width=\linewidth]{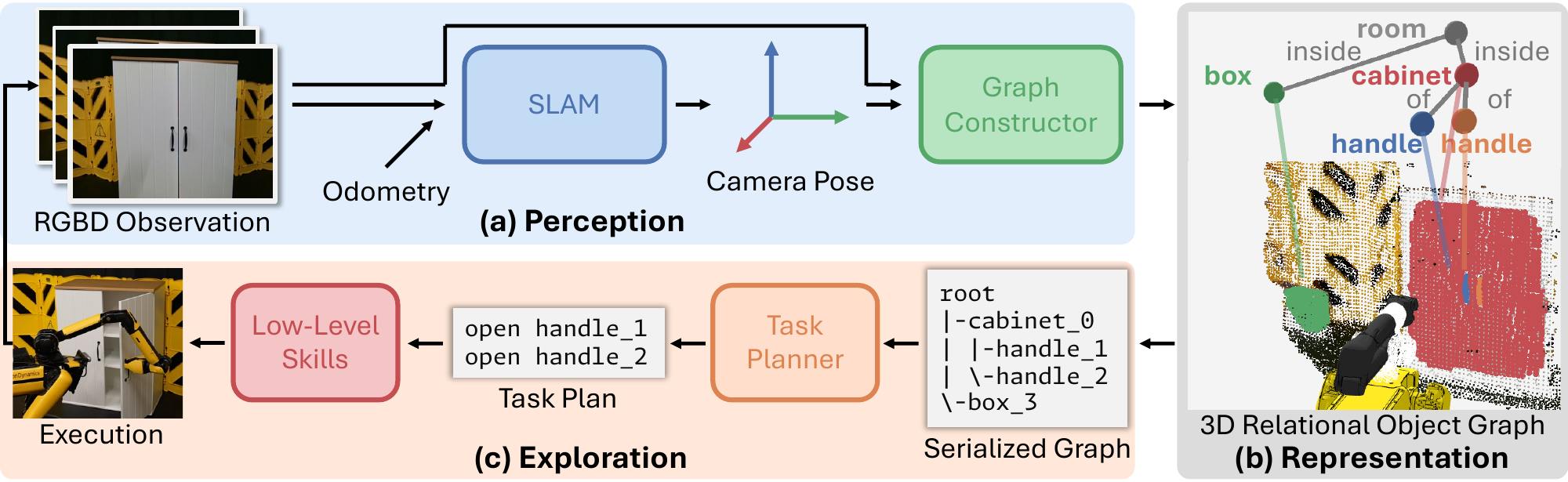}
    \vspace{-18pt}
    \captionof{figure}{\small
    \textbf{Method Overview.} (a) In the perception pipeline, SLAM processes RGBD observations and odometry estimation from the robot to output camera poses, which are used alongside the RGBD observations to construct an actionable 3D relational object graph. (b) The 3D relational object graph comprises object nodes containing both geometric and semantic information, as well as object edges that encode complex object relations. (c) The serialized object graph is fed into the task planner, and the generated task plans are executed using low-level skills to interactively explore the environment.
    }
    \label{fig:method}
    \vspace{-20pt}
\end{figure*}

\vspace{-5pt}
\subsection{3D Scene Graph for Robotics}
\vspace{-3pt}

3D scene graph representation is widely used in robotic manipulation and navigation
~\cite{maggio2024clio, jiang2024roboexp, jatavallabhula2023conceptfusion, peng2023openscene, ha2022semantic, yamazaki2024open}.
These representations often leverage 2D VFMs such as SAM, CLIP, or DINO~\cite{kirillov2023segany, pmlr-v139-radford21a, oquab2023dinov2} to extract 2D visual information, which is then fused into 3D space. However, existing methods tend to focus on the semantic understanding of objects, rather than encoding complex object relations such as \texttt{on}, \texttt{inside}, or \texttt{behind}. Understanding such occlusion relations is crucial for making informed decisions about where to explore and how to manipulate objects. In contrast, our representation encodes various types of occlusion relations in real-world environments, allowing the mobile robot to actively decide how to explore the environment. %
Although works such as ConceptGraph and SceneGPT~\cite{gu2024conceptgraphs, chandhok2024scenegpt} account for spatial relationships, they do not consider active interactions with the environment, such as opening drawers. In contrast, our representation considers how different actions can modify the environment (e.g., opening a drawer to retrieve a toy \texttt{inside}), allowing the system to choose the appropriate exploration and manipulation skills.

Additionally, prior works have proposed Action-Oriented Semantic Maps (AOSM) and similar 3D object-centric semantic representations for various tasks, including MR teleoperation~\cite{rosen2020building}, skill learning~\cite{rosen2023synthesizing}, and human-robot interaction~\cite{quesada2022proactive}. However, these approaches do not address interactive mobile exploration tasks, which require the scene representation to capture object relations, reason about unknown spaces, and decide the appropriate manipulation skills.

\vspace{-5pt}
\subsection{Foundational Model for Robotics}
\vspace{-3pt}

Many previous studies have used the generalization capabilities, common sense reasoning, and long-horizon planning abilities of VFMs and LLMs for robotic tasks such as manipulation~\cite{huang2023voxposer, driess2023palm,  liang2023code}, navigation~\cite{gu2024conceptgraphs, jatavallabhula2023conceptfusion}, and planning~\cite{yang2024llm, hu2023look}. However, these studies did not explore the potential of using VFMs and LLMs for active mobile exploration. In our work, we leverage VFMs~\cite{Cheng2024YOLOWorld, kirillov2023segany} to build 3D actionable relational object graphs. We then employ an LLM for decision-making based on an explicit 3D object graph representation of the environment, which our experiments demonstrate to be more efficient and effective than relying on memorizing 2D observation history~\cite{achiam2023gpt}.

\vspace{-5pt}
\section{METHOD}
\vspace{-5pt}

As shown in Figure~\ref{fig:method}, our framework consists of four modules - SLAM, Graph Constructor, Task Planner, and Low-Level Skills, each of which will be explained in detail in the following sections.

\vspace{-7pt}
\subsection{Problem Statement}
\label{sec:problem}
\vspace{-4pt}

\begin{figure*}
    \centering
    \includegraphics[width=\linewidth]{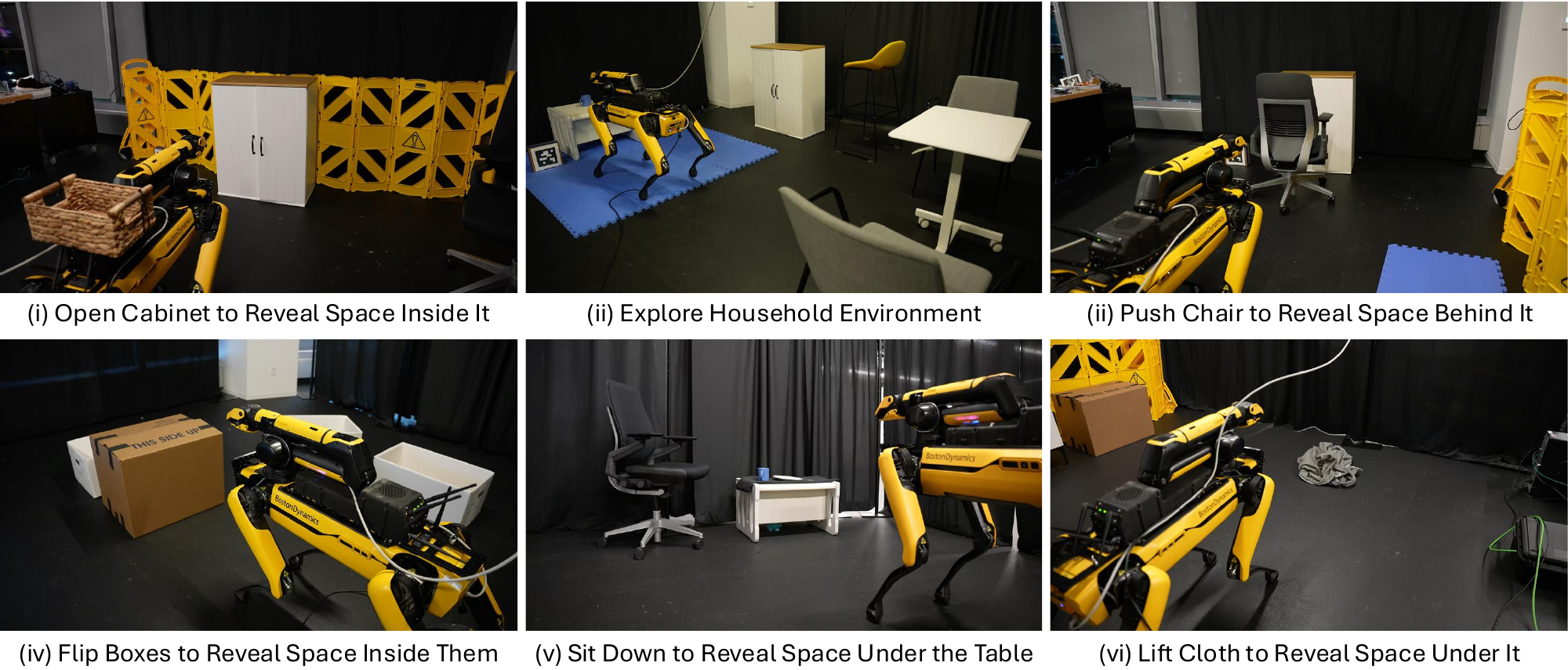}
    \vspace{-20pt}
    \captionof{figure}{\small
    \rebuttal{\textbf{Testing Scene Setup.} Our testing scenes cover diverse object instances, relations, and layouts to demonstrate our system's generalization capability across object instances, object relations and scenarios.}}
    \label{fig:all_room}
    \vspace{-18pt}
\end{figure*}

We define the object graph as $G=(V,E)$, where $V=\{v^0, v^1, \dots, v^N\}$ represents the set of object nodes, and $E=\{e^0, e^1, \dots, e^M\}$ represents the set of edges. Each node $v$ contains both semantic attributes, such as object labels, and geometric attributes, such as point clouds and normal estimations. Each edge $e$ represents a directed connection from node $v^i$ to node $v^j$, along with their object relationship. The objective of mobile exploration is to construct a graph $G$ that can minimize the unknown space, discover as many object nodes as possible, and establish correct object relationships.

\vspace{-7pt}
\subsection{SLAM}
\vspace{-4pt}

As shown in Figure~\ref{fig:method} (a), SLAM takes in a sequence of odometry estimation from robot and RGBD observations, $\mathbf{O}_{0...t}$, where each $\mathbf{O}_t \in \mathbb{R}^{H\times W\times 4}$ represents one RGBD frame, and  localizes the camera pose $\mathbf{T}_t$ at time $t$.
In practice, we use RTAB-Map v0.21.4 to estimate the camera pose~\cite{labbe2019rtab}.

\vspace{-7pt}
\subsection{Graph Constructor}
\vspace{-4pt}

Given the current RGBD observation $\mathbf{O}_t$, the corresponding camera pose $\mathbf{T}_t$, and the graph from the previous frame $G_{t-1}$, we construct the graph $G_t$ at time $t$ as illustrated in Figure~\ref{fig:method} (a) and (b). We first segment the objects using YOLO-World and SAM and obtain corresponding 3D point clouds~\cite{Cheng2024YOLOWorld, kirillov2023segany}. Next, we associate the segmented objects with previous object nodes based on geometric information and fuse the current observation to obtain the current object nodes. Finally, we establish object relationships by jointly considering geometric, semantic, and action-related information.

\rebuttal{\textbf{Object detection and association.}} We detect objects \rebuttal{with confidence higher than 0.1} and obtain the corresponding 3D point clouds $P_{t}=\{p_t^1, ..., p_t^K\}$, where $p_t^i$ is the point cloud of the $i^\text{th}$ object. We then associate point clouds at time $t$ with last frame object point clouds  $P_{t-1}=\{p_{t-1}^1, ..., p_{t-1}^N\}$. The association is resolved by checking detection label consistency and calculating the Intersection over Union (IoU) between $P_{t-1}$ and $P_t$.
Specifically, we create a value matrix $C \in \mathbb{R}^{K \times N}$, where each element is defined as follows:
\begin{align}
    C_{ij}=
    \begin{cases}
    \text{IoU}(p_t^i, p_{t-1}^j), & \text{if they have the same label} \\
    0, & \text{otherwise.}
    \end{cases}
\end{align}
For the $i^\text{th}$ detected object $p_t^i$, if $\max_{j \in \{1,...,N\}} C_{ij}$ is below threshold \rebuttal{0.15}, it is considered a newly detected object. Otherwise, the $i^\text{th}$ detected object is associated with the existing object that has the highest $C_{ij}$ value. After associating the current detection with the existing object graph, we could update the existing object nodes with the current observation.

\rebuttal{\textbf{Relation construction.}}
\rebuttal{We add edges using two complementary signals.}
\rebuttal{(i) \emph{Interaction-driven edges:} when an interactive skill reveals a new object in the new frame, we assign relations based on the last action (e.g., \texttt{open/flip}$\rightarrow$\texttt{inside}, \texttt{lift/check\_bottom}$\rightarrow$\texttt{under}, \texttt{push}$\rightarrow$\texttt{behind}).}
\rebuttal{(ii) \emph{Geometry-driven edges:} we use simple tests on 3D bounding boxes to construct geometric relations such as \texttt{on} edges.}

\rebuttal{\textbf{Voxel map.}}
\rebuttal{To represent unexplored and occluded space, we maintain a fixed 3D voxel grid and assign each voxel one of four labels: \texttt{unexplored}, \texttt{free}, \texttt{unknown}, or \texttt{outside}. Specifically, \texttt{unexplored} voxels have not been observed by the camera, \texttt{free} voxels lie in free space, \texttt{unknown} voxels are occluded behind a visible surface, and \texttt{outside} voxels fall outside the room boundaries.}

\rebuttal{We obtain voxel labels via depth testing. Concretely, we densely sample camera rays from the RGB-D sensor and compare ray depth to the measured depth: voxels in front of the first observed surface are labeled \texttt{free}, voxels behind it are labeled \texttt{unknown}, and rays with invalid depth remain \texttt{unexplored}. Voxels outside a predefined workspace box (or beyond wall/floor planes) are labeled \texttt{outside}.}

\rebuttal{This voxel map provides occlusion cues to decide whether the object is an obstruction (e.g., a large \texttt{unknown} region inside a closed cabinet suggests it obstructs the observation). The obstruction will be a semantic attribute for each node.}

\vspace{-7pt}
\subsection{Task Planner}
\vspace{-4pt}

We input the serialized object graph into the LLM to plan skills, as shown in Figure~\ref{fig:method} (c). For serialization, we perform a \rebuttal{standard depth-first search over the object graph to traverse all nodes in the graph and get the corresponding node depth levels. Then we serialize it to plain texts so that LLM can better understand the scene structure. If an object node is an obstruction, serialized text will append ``[obstruction]'' label to the node name.}

Specifically, we start at the root node and find all child nodes connected to it. We then put these nodes into a stack, which stores the nodes we plan to visit. Next, we pop the top node from the stack and check its children. If it has no children, it is a leaf node; otherwise, we repeat this operation until the stack is empty. Additionally, we keep track of information such as depth, object name, node ID, and related action details to build the serialized graph. Additionally, we provide the LLM with several prompts for planning. Our appendix \rebuttal{on \href{https://bdaiinstitute.github.io/curiousbot/}{website}} provides more details regarding the method, including search algorithm details and prompts. In our work, we choose GPT-4o as the LLM and it can be replaced with a stronger LLM in the future in a plug-and-play manner as needed, thanks to our modular design. Full prompts are provided in the appendix.

\begin{figure*}
    \centering
    \includegraphics[width=\linewidth]{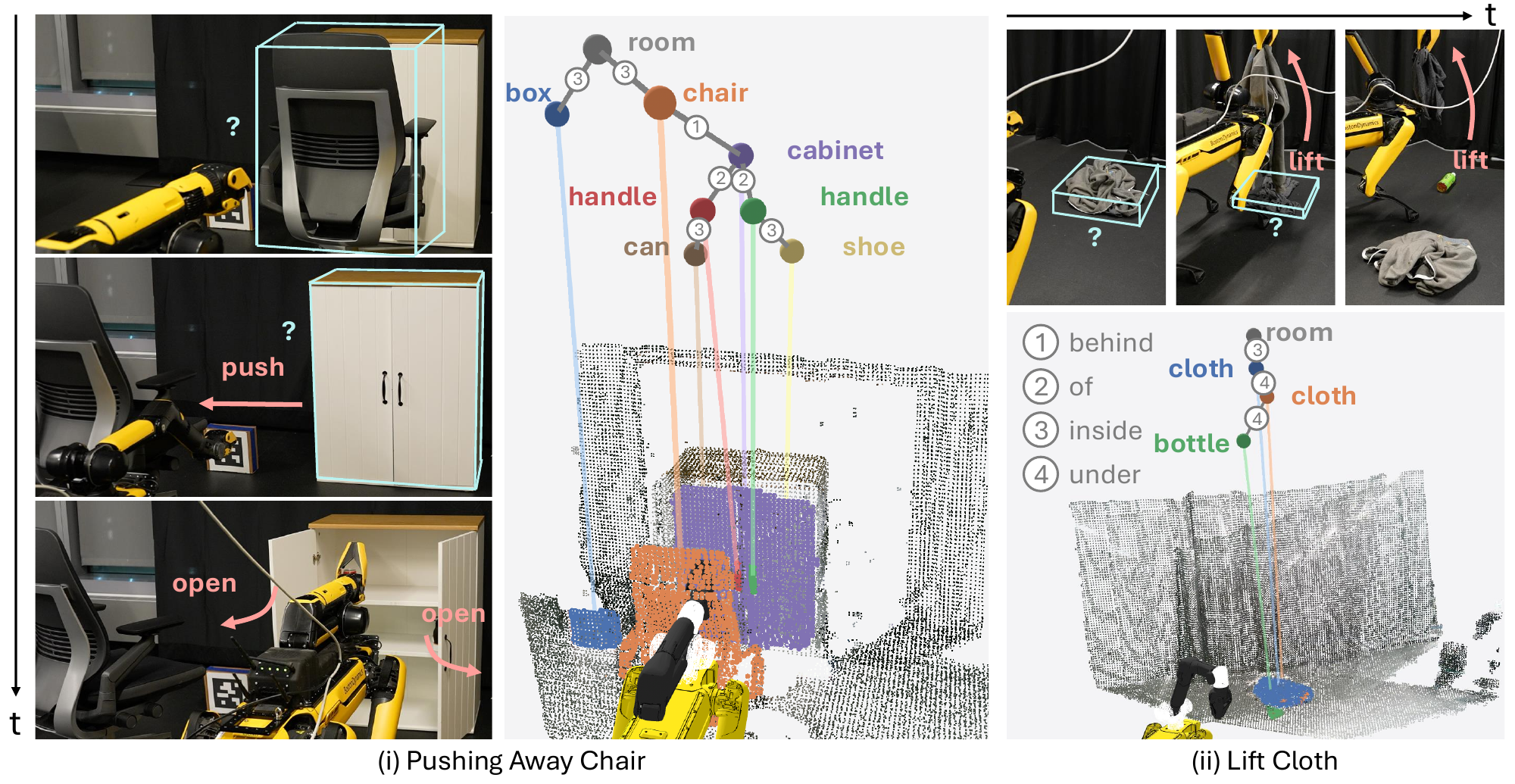}
    \vspace{-20pt}
    \captionof{figure}{\small
    \textbf{Diverse Scenes and Skills.} We evaluate our system's exploration capabilities across various tasks, including pushing the chair aside to reveal space behind it, lifting cloth to check underneath, flipping open boxes to inspect the contents, and exploring a household scene. These tasks showcase the system's ability to generalize across different object instances, scenarios, and object relations. \rebuttal{Note that question marks denote occluded regions inferred from voxel map instead of graph nodes}. Additional tasks can be found on the \href{https://bdaiinstitute.github.io/curiousbot/}{project page}.}
    \label{fig:all_exp_p1}
    \vspace{-18pt}
\end{figure*}

\vspace{-7pt}
\subsection{Low-Level Skills}
\vspace{-4pt}

We implement several primitive skills, including opening, lifting, pushing, collecting objects, sitting, and flipping. The skill output by the task planner consists of the skill name from our skill library and the target object index.
\rebuttal{These skills then query the 3D relational object graph and leverage the objects' semantic and geometric information to ground actions in the real world.}
Given this skill information, we execute the corresponding skill to explore the environment, as demonstrated in Figure~\ref{fig:method} (c).

Specifically, we construct the following manipulation skills using heuristics, including \texttt{open}, \texttt{flip}, \texttt{lift}, \texttt{push}, \texttt{sit}, and \texttt{collect}:
\begin{itemize}[leftmargin=10pt]
    \item \texttt{open}: We first detect the handle object node and the cabinet object node. Using Principal Component Analysis (PCA), we determine the axis of the handle and the normal axis of the cabinet. Then we can command the end-effector to approach the handle and grasp it. Since the cabinet’s articulation is unknown, we initially open the door using impedance control and adjust the target end-effector position and orientation according to manipulation feedback.

    In addition, to make the skill more robust, we take in the grasping feedback. If the grasping fails, we will retry the grasping with different action parameters.

    \item \texttt{flip}: We assume that the robot only flips boxes that are open. The end-effector first approaches the open box from above in a top-down pose, then pushes the box down on its side to flip it over.

    \item \texttt{lift}: We assume that the robot only lifts clothes. The end-effector first approaches the cloth and grasps its center of mass using the top-down pose.

    \item \texttt{push}: In our experiments, the robot will push large objects aside. It will first walk to the front of the object, then move aside. To push the object, it will extend its arm and move sideways while maintaining the arm in an extended position.

    \item \texttt{sit}: In our experiments, the robot will walk to the front of the object and then sit down to check the space beneath it. This sitting action can be executed using the Spot API.

    \item \texttt{collect}: To collect the object, we will first use an off-the-shelf grasping planner or heuristics to grasp it. Then, we command the end-effector to move to the designated position and place the object onto the blanket.
\end{itemize}

\vspace{-10pt}

\subsection{\rebuttal{Graph Updates}}
\vspace{-4pt}

\rebuttal{After executing the low-level skills, we update the serialized scene graph and no longer label the explored object node as an obstruction, informing the task planner whether an object is explored and avoiding repeatedly revisiting the same object.}

\begin{figure}[!ht]
    \centering
    \vspace{-10pt}
    \includegraphics[width=\linewidth]{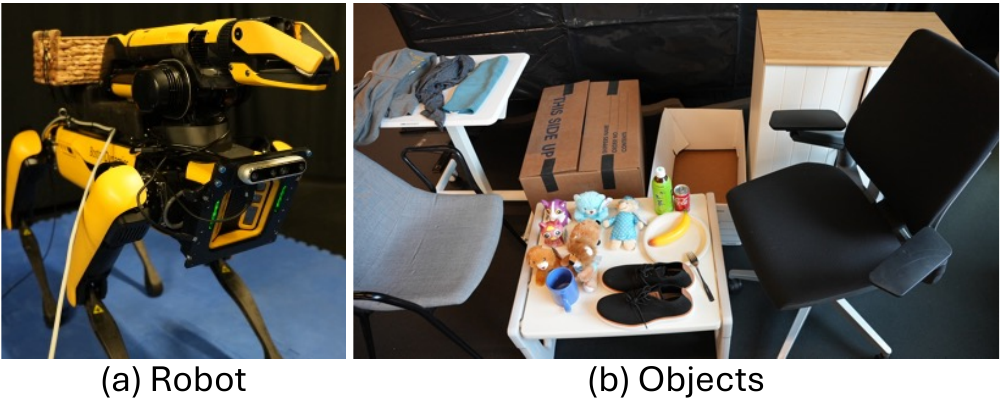}
    \vspace{-18pt}
    \captionof{figure}{\small
    \textbf{Experiment Setup.} (a) illustrates the use of a Spot robot equipped with an external RealSense 455. (b) showcases the diverse objects used, emphasizing the system's generalization capabilities across various object \rebuttal{instances}.
    }
    \label{fig:setup}
    \vspace{-10pt}
\end{figure}

\section{EXPERIMENT}
\vspace{-5pt}

In our experiments, we aim to answer the following questions:
(1) What kinds of tasks can be enabled by our system, and what scenarios can our robot explore?
(2) How does each component perform, and what are the common failure patterns?
(3) How will the whole system perform with different design choices?

\vspace{-8pt}
\subsection{Experiment Setup}
\vspace{-3pt}

We use the Boston Dynamics Spot and Spot API v4.0.0 for low-level control. An additional RealSense 455 camera is installed at the front to enhance environmental observation, as shown in Figure~\ref{fig:setup}.
For computation, we use a desktop equipped with an Nvidia RTX A6000 GPU and an AMD CPU with 128GB of memory.
\rebuttal{The graph construction can run around 3Hz given this hardware.}
The robot is connected to this machine using Ubuntu 22.04 through Ethernet. Our system is evaluated on diverse daily objects, as shown in Figure~\ref{fig:setup}. We set up the environment in a 3m $\times$ 4m room.
\camready{We use 12 object categories and 6 unique room layouts for our testing.}

\vspace{-8pt}
\subsection{Mobile Exploration in Various Scenes}
\vspace{-3pt}

\begin{figure*}
    \centering
    \includegraphics[width=\linewidth]{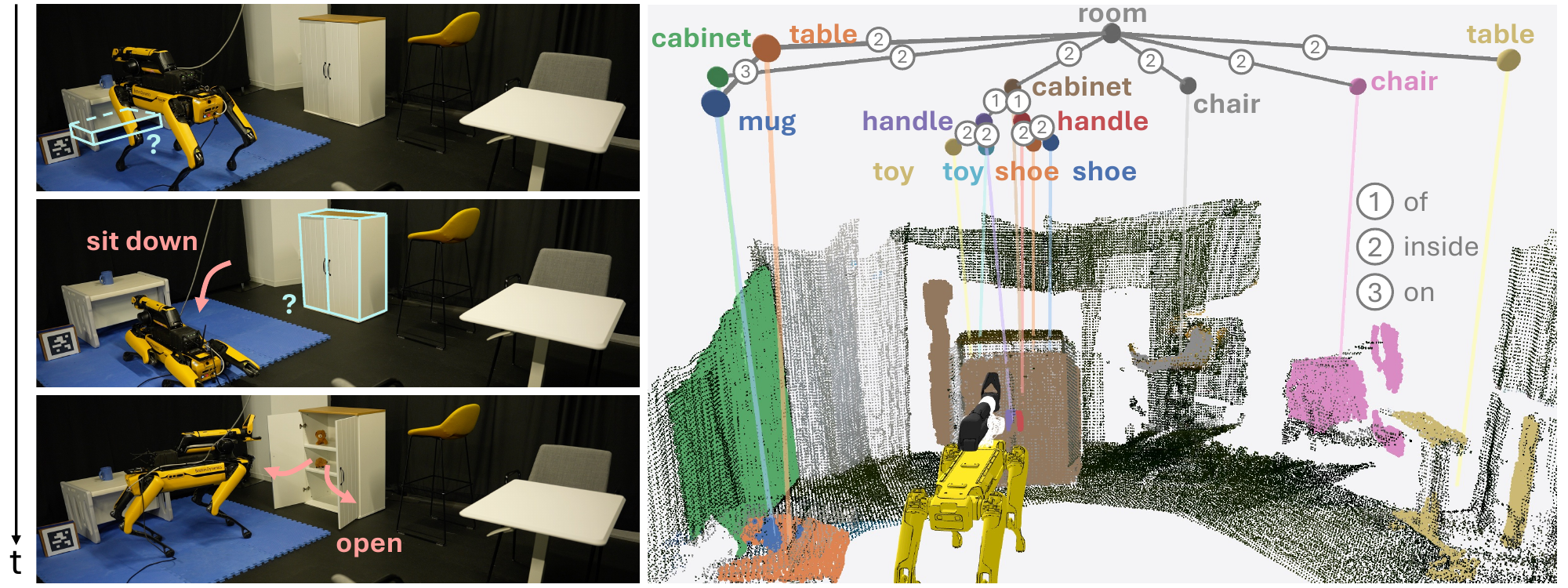}
    \vspace{-15pt}
    \captionof{figure}{\small
    \textbf{Household Scene.} We build a household environment and evaluate our system performance. Our system successfully builds an object graph covering all objects and relations. In addition, it sits down and opens the cabinet to actively reveal the unknown space.}
    \label{fig:all_exp_p2}
    \vspace{-18pt}
\end{figure*}

We qualitatively evaluate our system on diverse scenes, as shown in Figure~\ref{fig:all_exp_p1} and Figure~\ref{fig:all_exp_p2}. We would like to highlight the following aspects of our system's capabilities:

\textbf{Diverse Object Categories.} Our system operates in scenarios containing various types of objects, such as articulated objects, deformable objects, and rigid objects, demonstrating its generalization capabilities across different object instances.

\textbf{Various Object Relations.} Our system encodes five types of object relations commonly observed in the real world, including \texttt{behind}, \texttt{of}, \texttt{inside}, \texttt{on}, and \texttt{under}. For instance, the robot understands that there is unknown space behind the chair, requiring it to push the chair away to reveal the space behind, as shown in Figure~\ref{fig:all_exp_p1}.

\textbf{Different Layouts.} Testing scenarios include scenes of varying scales and layouts, from piles of cloth to larger household environments, which demonstrate that our system can \rebuttal{work for} various scenes.

\textbf{Diverse Interactions.} The robot interacts with the environment and explores the scene in multiple ways. For instance, the robot can grasp an object rigidly, such as opening a cabinet, and also interact with objects nonprehensilely using its arm instead of the gripper, such as pushing the chair, as shown in Figure~\ref{fig:all_exp_p1}. Additionally, the robot can actively move the camera around without manipulating objects, such as sitting down to check the space under a table, as shown in Figure~\ref{fig:all_exp_p2}.

\vspace{-5pt}

\begin{figure}
    \centering
    \includegraphics[width=\linewidth]{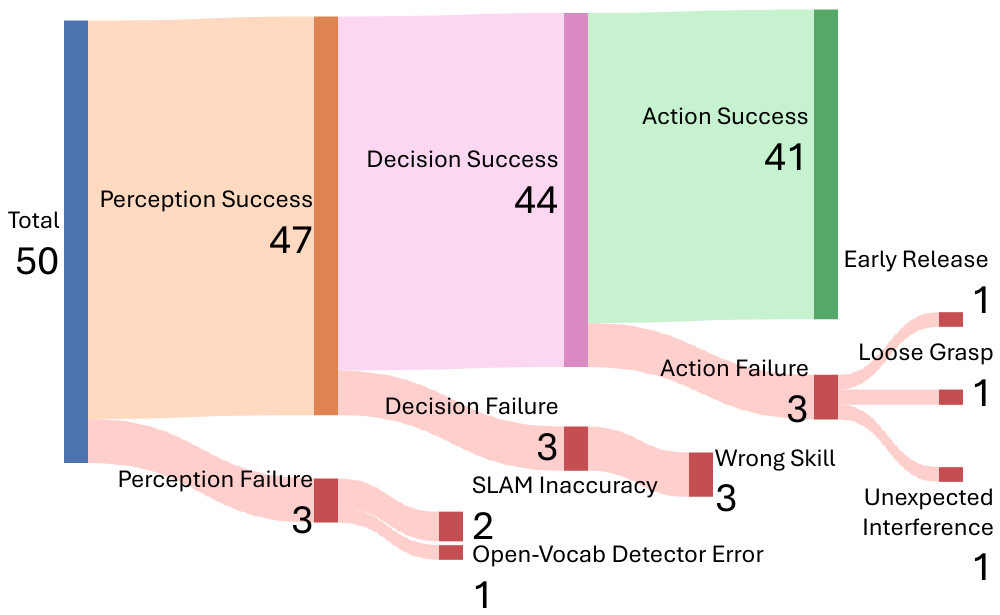}
    \vspace{-15pt}
    \captionof{figure}{\small
    \textbf{Failure Breakdown.} We analyze the failure modes of our system during exploration tasks, identifying three main causes: perception failure, decision failure, and action failure.
    }
    \label{fig:error}
    \vspace{-16pt}
\end{figure}

\vspace{-5pt}
\subsection{Failure Breakdown}
\label{sec:breakdown}
\vspace{-3pt}

We conduct experiments involving tasks such as flipping boxes, opening drawers, checking underneath objects, pushing boxes, and lifting cloth. Each task is repeated ten times under different initial conditions, and Figure~\ref{fig:error} shows the failure breakdown of all evaluation rollouts. A rollout is considered successful if the robot completes all exploration skills.

The overall success rate is 82\%. For the failure cases, we categorize the primary reasons into perception failure, decision failure, and action failure. Perception failure occurs when inaccurate perception results lead to unreasonable plans in downstream task planning or incorrect action execution. Decision failure happens when, despite having a correct serialized graph, the task planner makes an incorrect decision. Action failure occurs when, despite having a correct task plan and an accurate object graph, the skill execution fails.

In cases of perception failure, two major causes are an inaccurate object graph due to imprecise SLAM and errors from the open-vocabulary object detector. For decision failure, the task planner can fail in predicting the correct skills for the corresponding object nodes. Regarding action failures, we highlight the complexity of real-world manipulation, such as the early release of the gripper, loose grasping, and unexpected interference between the robot and the object.

\vspace{-5pt}
\subsection{Comparisons with Baselines}
\label{sec:comp_with_baselines}
\vspace{-3pt}

We compare our system with the following VLM baselines on the same five tasks as in Section~\ref{sec:breakdown} to test the hypothesis: is reasoning about interactive mobile exploration tasks more effective in 2D or 3D? For the 2D VLM baselines, we assume perfect low-level execution, making them stronger.

\begin{itemize}[leftmargin=10pt]
    \item \textbf{LLaVa:} We directly feed the current RGB observation and the same text prompt as our LLM task planner into LLaVa, the state-of-the-art open-source VLM~\cite{liu2023improvedllava}. Because VLM does not equip manipulation skills, a human operator will help VLM finish manipulation.
    \item \textbf{Gemini:} Similar to the LLaVa baseline, with the only change being the substitution of LLaVa with Gemini, a state-of-the-art closed-source VLM~\cite{team2023gemini}.
    \item \textbf{GPT-4o:} Similarly, we replace the VLM with GPT-4o, another state-of-the-art VLM~\cite{achiam2023gpt}.
    \item \textbf{Heuristics:} We implement a heuristic exploration policy where the robot will open all handles.
\end{itemize}

\vspace{-8pt}

\begin{table}[htbp]
\centering
\small
\begin{tabular}{lccccc}
\toprule
 & LLaVa & Gemini & GPT-4o & Heuristics & Ours \\
\midrule
\multicolumn{6}{c}{\textbf{Flipping Boxes}}\\
\midrule
Success $\uparrow$      & 0\%          & 0\%          & 0\%          & 0\%          & \textbf{80\%} \\
OR $\uparrow$   & 0\%          & 0\%          & 0\%          & 0\%          & \textbf{70\%} \\
GED $\downarrow$             & 2.3          & 2.3          & 2.1          & 2            & \textbf{1}    \\
\midrule
\multicolumn{6}{c}{\textbf{Opening Drawers}}\\
\midrule
Success $\uparrow$      & 40\%         & \textbf{80\%}& 60\%         & 60\%         & \textbf{80\%} \\
OR $\uparrow$   & 60\%         & \textbf{90\%}& 80\%         & 72\%         & 88\%          \\
GED $\downarrow$             & 7.5          & 5.9          & 6.1          & 3            & \textbf{2.4}  \\
\midrule
\multicolumn{6}{c}{\textbf{Checking Underneath}}\\
\midrule
Success $\uparrow$      & 60\%         & 40\%         & 0\%          & 0\%          & \textbf{90\%} \\
OR $\uparrow$   & 60\%         & 40\%         & 0\%          & 0\%          & \textbf{90\%} \\
GED $\downarrow$             & 2.2          & 2.7          & 3.2          & 3.1          & \textbf{1.5}  \\
\midrule
\multicolumn{6}{c}{\textbf{Pushing Boxes}}\\
\midrule
Success $\uparrow$      & 0\%          & 0\%          & 0\%          & 0\%          & \textbf{70\%} \\
OR $\uparrow$   & 0\%          & 0\%          & 0\%          & 0\%          & \textbf{70\%} \\
GED $\downarrow$             & 4.1          & 4.3          & 4            & 4            & \textbf{1.2}  \\
\midrule
\multicolumn{6}{c}{\textbf{Lifting Cloth}}\\
\midrule
Success $\uparrow$      & 10\%         & 40\%         & \textbf{100\%}& 0\%         & 90\%          \\
OR $\uparrow$   & 10\%         & 40\%         & \textbf{100\%}& 0\%         & 90\%          \\
GED $\downarrow$             & 2            & 1.9          & 1.2          & 2.2          & \textbf{0.3}  \\
\midrule
\multicolumn{6}{c}{\textbf{Average}}\\
\midrule
Success $\uparrow$      & 22\%         & 32\%         & 32\%         & 12\%         & \textbf{82\%} \\
OR $\uparrow$   & 26\%         & 34\%         & 36\%         & 14.4\%       & \textbf{81.6\%} \\
GED $\downarrow$             & 3.62         & 3.42         & 3.32         & 2.86         & \textbf{1.28} \\
\bottomrule
\end{tabular}
\vspace{-5pt}
\caption{\small \textbf{Quantitative Results.} We quantitatively evaluate our system on five tasks, each repeated ten times, and compare it with four baselines: LLaVa, Gemini, GPT-4o, and heuristics. The evaluation metrics include success rate, Object Recovery (OR), and Graph Editing Distance (GED), and metrics are defined in Section~\ref{sec:comp_with_baselines}. Our results show that our approach is more effective at accomplishing exploration tasks and capable of constructing accurate object graphs.}
\vspace{-15pt}
\label{tab:quant}
\end{table}

We evaluate performance using three metrics: 1) \textbf{Success Rate}: A rollout is considered successful if all exploration skills are correctly executed. 2) \textbf{Object Recovery (OR)}: Assuming the ground truth object nodes are $V_{\text{gt}}$ and the discovered object nodes are $V$, object discovery is defined as $|V_{\text{gt}} \cap V| / |V_{\text{gt}}|$. 3) \textbf{Graph Editing Distance (GED)}: If the cost of adding, deleting, or moving one edge or node is 1, GED is defined as the total cost of editing the final graph $G$ to match the ground truth graph $G_{\text{gt}}$.

\begin{table}[htbp]
\small
\centering
\begin{tabular}{ccccc}
\toprule
Number of Examples & 7 (Ours) & 5 & 3 & 1 \\
\midrule
Success Rate & \textbf{89\%} & 67\% & 56\% & 11\% \\
Object Recovery & \textbf{89\%} & 67\% & 56\% & 11\% \\
GED & \textbf{0.33} & 0.89 & 1.00 & 2.67 \\
\bottomrule
\end{tabular}
\caption{\small \textbf{Ablation Study.} We examine how our system's performance changes based on the number of examples fed into the LLM. This figure shows that performance worsens with less examples. Meanwhile, given the current number of examples, our system achieves a high success rate, demonstrating that the provided examples are both minimal and necessary.}
\vspace{-18pt}
\label{tab:ablation}
\end{table}

Table~\ref{tab:quant} summarizes our quantitative results. We found our 3D relational object graph is more effective than feeding RGB observations into a VLM. This is because our representation explicitly reason the topological relationships of object nodes, leading to more effective task planning compared to requiring VLM to memorize observations and reason object relations implicitly. Additionally, our actionable object graph grounds actions in the representation, while RGB observations alone do not provide sufficient information for low-level skill selection. Additionally, while simple exploration heuristics may yield comparable performance in certain tasks, they do not work for other tasks.

\vspace{-5pt}
\subsection{Ablation Study}
\vspace{-3pt}

We also study how our system's performance varies with the number of examples provided to the LLM. We reduce the number of examples from 7 to 1 and evaluate the performance on three tasks: flipping boxes, pushing boxes, and lifting cloth, with each task repeated three times. Table~\ref{tab:ablation} shows performance decreases as the number of examples decreases, underscoring that the examples we provided to the LLM are both minimal and necessary for task planning.

\vspace{-5pt}

\section{CONCLUSION}

\vspace{-3pt}

Interactive mobile exploration has been a longstanding and essential problem in robotics. However, existing approaches to mobile exploration primarily focus on active perception rather than active interaction, which limits the robot's ability to fully explore the environment. Current methods for robotic exploration via active interaction are mainly focused on tabletop scenes, overlooking the unique challenges of mobile settings, such as expansive exploration spaces, large action spaces, and diverse object relations. We introduce the 3D relational object graph, which encodes diverse object relations, and build a system capable of exploration through active interaction based on this representation. We evaluate our system in diverse scenes, demonstrating its effectiveness and generalization capabilities qualitatively. Our quantitative results further underscore its effectiveness compared to directly using VLMs.

\ralparagraph{\rebuttal{Limitation}}
\rebuttal{While our work incorporates diverse manipulation skills for exploration, skill acquisition currently requires experienced roboticists to write and fine-tune the heuristics through trial and error.}
\rebuttal{These skills might fail if involving heavily occluded scenes.}
\rebuttal{Developing a scalable skill acquisition process would equip our robot with a broader range of skills and enable more diverse exploration.}
\rebuttal{Additionally, since maintaining dynamic scene memory remains an open challenge, we do not update the 3D scene graph after each interaction. In future work, developing a dynamic scene memory that can track objects over time would be highly valuable.}

\rebuttal{While this work captures various object relations commonly seen in the real world and sufficient for diverse interactive exploration tasks, the real world relations can be even more complex, such as \texttt{next to}. A promising future direction will be capturing object relations automatically using foundation models and develop more expressive representation.}

\bibliographystyle{IEEEtran}
\bibliography{references}

\end{document}